\documentclass[letterpaper, 10 pt, journal, twoside]{ieeetran}
\usepackage{amsmath,amsfonts}
\usepackage{algorithmic}
\usepackage{array}
\usepackage[export]{adjustbox}
\usepackage[caption=false,font=normalsize,labelfont=sf,textfont=sf]{subfig}
\usepackage{textcomp}
\usepackage{stfloats}
\usepackage{url}
\usepackage{verbatim}
\usepackage{graphicx}

\usepackage{times}
\usepackage{multicol}
\usepackage{hyperref}
\usepackage{graphicx}
\usepackage{bm}
\usepackage[nolist]{acronym}
\usepackage{amsmath}
\usepackage{amssymb}
\usepackage[capitalise]{cleveref}
\usepackage{algorithm}
\usepackage[usenames,dvipsnames,table]{xcolor}
\usepackage{algorithmic}
\usepackage{booktabs}
\usepackage{array}
\usepackage{tabularx}
\usepackage{enumitem}
\usepackage{caption}
\usepackage{diagbox}
\usepackage{etoolbox}\AtBeginEnvironment{algorithmic}{\small}

\newcommand{\specialcell}[2][l]{%
  \begin{tabular}[#1]{@{}l@{}}#2\end{tabular}}

\DeclareMathAlphabet{\pazocal}{OMS}{zplm}{m}{n}
\DeclareMathOperator{\E}{\mathbb{E}}

\usepackage{etoolbox}
\apptocmd{\thebibliography}{\scriptsize}{}{}

\def\BibTeX{{\rm B\kern-.05em{\sc i\kern-.025em b}\kern-.08em
    T\kern-.1667em\lower.7ex\hbox{E}\kern-.125emX}}
\usepackage{balance}

\begin{acronym}
\acro{MDP}{Markov Decision Process}
\acro{NN}{Neural Network}
\acro{DL}{Deep Learning}
\acro{IL}{Imitation Learning}
\acro{MPC}{Model Predictive Control}
\acro{MPPI}{Model Predictive Path Integral}
\acro{BC}{Behavior Cloning}
\acro{RL}{Reinforcement Learning}
\acro{IRL}{Inverse Reinforcement Learning}
\acro{OC}{Optimal Control}
\acro{IOC}{Inverse Optimal Control}
\acro{MEDIRL}{Maximum Entropy Deep IRL}
\acro{SVF}{State Visitation Frequency}
\acro{SVFs}{State Visitation Frequencies}
\acro{VI}{Value Iteration}
\acro{PPO}{Proximal Policy Optimization}
\acro{QP}{Quadratic Programming}
\acro{ADE}{Average Displacement Error}
\end{acronym}

\usepackage{fancyhdr}

\begin{document}
\title{Spatiotemporal Costmap Inference for MPC\\via Deep Inverse Reinforcement Learning}

\markboth{IEEE Robotics and Automation Letters. Preprint Version. Accepted January, 2022}
{}

\author{Keuntaek Lee$^{1}$, David Isele$^{2}$, Evangelos A. Theodorou$^{1}$, and Sangjae Bae$^{2}$%
\thanks{Manuscript received: September 7, 2021; Revised December 7, 2021; Accepted January 5, 2022.}
\thanks{This paper was recommended for publication by Editor Dan Popa upon evaluation of the Associate Editor and Reviewers’ comments.}
\thanks{This work was supported by Honda Research Institute USA, Inc.}
\thanks{$^{1}$Keuntaek Lee and Evangelos A. Theodorou are with the Georgia Institute of Technology, Atlanta, GA, USA.
        {\tt\footnotesize $\{$keuntaek.lee, evangelos.theodorou$\}$@gatech.edu}}
\thanks{$^{2}$David Isele and Sangjae Bae are with Honda Research Institute USA, Inc.
        {\tt\footnotesize $\{$disele,sbae$\}$@honda-ri.com}}
}

\maketitle
\thispagestyle{fancy}

\begin{abstract}
It can be difficult to autonomously produce driver behavior so that it appears natural to other traffic participants. Through Inverse Reinforcement Learning (IRL), we can automate this process by learning the underlying reward function from human demonstrations. We propose a new IRL algorithm that learns a goal-conditioned spatio-temporal reward function. The resulting costmap is used by Model Predictive Controllers (MPCs) to perform a task without any hand-designing or hand-tuning of the cost function. We evaluate our proposed Goal-conditioned SpatioTemporal Zeroing Maximum Entropy Deep IRL (GSTZ)-MEDIRL framework together with MPC in the CARLA simulator for autonomous driving, lane keeping, and lane changing tasks in a challenging dense traffic highway scenario. Our proposed methods show higher success rates compared to other baseline methods including behavior cloning, state-of-the-art RL policies, and MPC with a learning-based behavior prediction model.
\\ Supplementary video: \textcolor{blue}{\url{https://youtu.be/Ft4u5AclSLA}}
\end{abstract}

\begin{IEEEkeywords}
Learning from Demonstration, Reinforcement Learning, Optimization and Optimal Control, Motion and Path Planning, Autonomous Vehicle Navigation
\end{IEEEkeywords}

\section{Introduction}
\IEEEPARstart{O}{bjective} functions for autonomous driving often require balancing safety, efficiency, and smoothness amongst other concerns \cite{bae2019cooperation,iteKQ,isele2019itsc}.
While formulating such an objective is often non-trivial, the final result can produce behaviors that are unusual and difficult to interpret for other traffic participants, which in turn, can have an impact on the stated objective of safety \cite{schwall2020waymo}.
Instead of trying to adjust the weights or add additional objectives to an \ac{OC} problem to produce policies that appear more natural, 
we target natural behavior directly by use of a learning algorithm that learns a cost function from data.
In this paper, we specifically focus on learning the cost function of driving a vehicle on a highway, from human demonstrations.
A motivation of this work is to learn a cost function that any optimal controller can directly use to accomplish a task of autonomous driving on a highway, without any further information.

In autonomous driving, although we can have a near-perfect estimation of the other vehicles' state with various types of sensors, the challenging part in planning is to predict the other vehicles future states, as they are not static obstacles. Modeling the other vehicles' behavior and evaluating current and future safety related to them is a difficult problem. In this work, as human drivers make nearly optimal decisions considering the other vehicles' future states and avoiding future collisions, we aim to learn those implicitly in the form of a cost function, from human demonstrations by making an assumption that the driving demonstrations by human drivers show optimal behavior at an expert-level.

Given demonstrations of experts accomplishing a specific task, 
\ac{IOC} or \ac{IRL} can learn a cost function that explains the experts' demonstrated behavior.
\ac{OC}/\ac{RL} can then find an optimal policy that generates trajectories similar to that of the experts.
As shown in the literature \cite{Ng2000irl, abbeel2004apprenticeship}, the learned reward or cost function through \ac{IRL} can better generalize than state-to-state or state-to-action learning approaches like \ac{BC} \cite{pan2019imitation} and it provides an additional interpretable layer which can be used for debugging and verification.

Recently, a number of reward/cost function learning algorithms were introduced, including the Maximum Entropy \ac{IRL} (MaxEnt IRL) \cite{ziebart2008maximum}, Generative Adversarial Imitation Learning (GAIL) \cite{ho2016gail}, Guided Cost Learning (GCL) \cite{finn2016gcl}, Adversarial \ac{IRL} (AIRL) \cite{fu2018learning}, and the \ac{MEDIRL} \cite{wulfmeier2015medirl}.
They all showed great performance on learning a cost function from data, however, from the fact that the intuitive explainability is essential with deep \acp{NN}, in our work, we focus on representing the cost function as an image (map) \cite{drews2017costmap, lee2021approximateIRL} through \ac{MEDIRL}, to provide a quick and intuitive analysis for both humans and real-time OC/RL policies.

Our contributions are as follows:
\begin{itemize}
    \item {We propose an MPC framework that leverages \ac{IRL} to automate the cost and safety evaluations in autonomous driving.}
    \item {Our proposed Goal-conditioned SpatioTemporal Zeroing (GSTZ)-\ac{MEDIRL} framework improves the \ac{MEDIRL} algorithm and enhances the interpretability of a predicted costmap} that MPC can directly use without any further designing or tuning of the cost function.
    \item {With the costmap learned from human demonstrations}, we demonstrate successful autonomous driving, lane keeping, and lane changing in a dense traffic highway scenario with a \emph{real-time} MPC in CARLA simulator.
\end{itemize}

To the best of our knowledge, our work shows the first demonstration of real-time optimal control with the costmap learned through \ac{MEDIRL}.

\section{Related work}
\label{sec:related_work}

Our proposed cost function learning algorithm is built on top of the original \ac{MEDIRL} algorithm \cite{wulfmeier2015medirl, wulfmeier2016watchthis}.
{The authors} extended the original work \cite{wulfmeier2016watchthis} to learn costmaps for driving in complex urban environments and{, furthermore, they} addressed real-world challenges of applying the \ac{MEDIRL} algorithm in the urban driving environment \cite{wulfmeier2017largescale}.

In the same context of autonomous driving, the kinematics-integrated \ac{MEDIRL} \cite{zhang2018integrating} improved the performance of the \ac{MEDIRL} by adding extra information, the kinematics, and environmental context, to the algorithm.
Another similar work \cite{Jung2021MEDIRL} improved the \ac{MEDIRL} by adding multiple contexts; the history of observation information, the past trajectory, and the route plan.

Previous work using \ac{MEDIRL} algorithm \cite{wulfmeier2015medirl, wulfmeier2016watchthis, wulfmeier2017largescale, zhang2018integrating, Jung2021MEDIRL} used the \ac{VI} policy learning algorithms to solve the forward \ac{RL} problem inside the Inverse \ac{RL} loop, which could not be used at test time because of the two major problems:
\begin{itemize}
    \item The limitation of the real-time computation of the \ac{VI} algorithm.
    \item The unrealistic discrete action space. For example, 4 actions (up, down, left, right) \cite{zhang2018integrating} or 6 actions (stay, east, west, north, northwest, and northeast) \cite{Jung2021MEDIRL}.
\end{itemize}

The policy with a discrete action space (4-6) is not only infeasible for real-world autonomous driving, but also not suitable to be used as an optimal policy for learning a costmap if the costmap has a higher resolution, or the velocity of the demonstration is high.

This problem of using a \ac{VI} type policy learning motivated us to use a continuous state-action space policy for solving a forward \ac{RL} problem, which will be {detailed} in \cref{sec:optimal_control}.
Also, a costmap learned from these approaches cannot be used by MPC to accomplish a task without extra cost terms, e.g. a velocity cost, and this motivated our spatiotemporal costmap learning, which will be discussed in \cref{sec:spatiotemporal}.

Recently, in the learning-based autonomous driving literature, RL policies have shown great performance in autonomous driving in a dense traffic scenario. PPUU \cite{ppuu} learned a policy together with the world (forward) model in a stochastic fashion to predict and plan under uncertainty and DRLD \cite{Saxena2020RL} learned a policy with PPO \cite{ppo} algorithm. Although these state-of-the-art RL works solve a different problem but are highly related to our work as they use very similar observation information, a rasterized image input, for their NNs. We test these methods and report a comparative analysis in \cref{sec:experiments}.

\section{Preliminaries}
\subsection{\acl{IRL}}

In a typical \acp{MDP}, we have a 5-tuple, $(S,A,P_{sa},R,\gamma)$, where $S$ is a set of states $s$, $A$ is a set of actions $a$, $P_{sa}$ is a set of state transition probabilities, $R$ is the expected immediate reward received after transitioning from a current state to a next state, and $\gamma$ is the future discount factor for reward. \ac{IRL} aims to infer $R$ from a set of $N$ expert demonstrations $D=(\zeta_1, \zeta_2, ..., \zeta_N)$, where 
$\zeta_i=\{s_1,...,s_T\}$, with $T$ being the length/timesteps of a demonstration.

Ng and Russell \cite{Ng2000irl} introduced a feature-based linear reward function setting where the reward $R$ is parameterized by $\theta$: $R_\theta(s)=\theta^\top f(s),$ where $\theta \in \mathbb{R}^n$ is the weight parameter and $f(s): S\rightarrow \mathbb{R}^n$ represents state features.

Given the discount factor $\gamma$ and the policy $\pi$, the reward $R$ is the expected cumulative discounted sum of future reward:

\vspace{-0.75\baselineskip}
\small
\begin{align}
    \E\Big[\sum_{t=0}^\infty\gamma^t R_\theta(s_t)|\pi\Big] &= \E\Big[\sum_{t=0}^\infty\gamma^t \theta^\top f(s_t)|\pi\Big]
    = \theta^\top \bar{f}(\pi)
\end{align}
\normalsize
where $\bar{f}$ is defined as the expected cumulative discounted sum of feature values or feature expectations \cite{abbeel2004apprenticeship}.
Abbeel and Ng \cite{abbeel2004apprenticeship} showed that if the expert's and learner's feature expectations match, then the learner policy is guaranteed to perform as well as the expert policy.

\subsection{Maximum Entropy Inverse Reinforcement Learning}

Given the expert's demonstrations, if the expert's behavior is suboptimal (imperfect or noisy), it is hard to represent the behavior with a single reward function. Ziebart et al. \cite{ziebart2008maximum} introduced the Maximum Entropy \ac{IRL} approach to solve this ambiguity problem.

Maximizing the entropy of distributions over paths while satisfying the feature expectation matching constraints \cite{abbeel2004apprenticeship} is equivalent to maximizing the likelihood of the observed data $D$ under the assumed maximum entropy distribution \cite{jaynes1957}:

\vspace{-0.75\baselineskip}
\small
\begin{align}
\theta^* = \underset{\theta}{\text{argmax}} L(\theta) = \underset{\theta}{\text{argmax}} \sum_{\zeta \in D} \text{log} P(\zeta|\theta,P_{sa})
\end{align}
\normalsize
where $P(\zeta|\theta, P_{sa})$ follows the maximum entropy (Boltzmann) distribution \cite{ziebart2008maximum}.
This convex problem is solved by gradient-based optimization methods with 

\vspace{-0.75\baselineskip}
\small
\begin{align}
    \cfrac{\partial L(\theta)}{\partial \theta}
    &= \sum_{s \in \zeta \in D}\mu_s f(s) - \sum_{s_i}\mu_{s_i}f(s_i) \label{eq:meirl_update_original}
\end{align}
\normalsize
where $\mu_s$ is defined as the \ac{SVF}, the discounted sum of probabilities of visiting a state $s$:
$\mu_s = \sum_{t=0}^\infty \gamma^t P(s_t=s|\pi,\theta,P_{sa})$.
With a given or selected $f$, this update rule ends up as finding $\theta$ in reward that an optimal policy matches the SVF of the demonstration $D$.

\subsection{Maximum Entropy Deep Inverse Reinforcement Learning}

Previous approaches to estimate a reward function used a weighted linear reward function with hand-selected features. To overcome the limits of the linear expression,
Wulfmeier et al. \cite{wulfmeier2015medirl} introduced using \acp{NN} to extend the linear reward to nonlinear reward, $R_\theta(s)=R(f(s),\theta)$.
By training a \ac{NN} with a raw observation obtained from sensors as an input, both the weight and the features are automatically obtained, so it does not require hand-designing state features.

In \ac{MEDIRL}, the network is trained to maximize the joint probability of the demonstration data $D$ and model parameters $\theta$ under the estimated reward $R_\theta(s)$:

\vspace{-0.75\baselineskip}
\small
\begin{align}
    L(\theta) &= \text{log}P(D,\theta | R_\theta(s)) \\
    &= \text{log}P(D|R_\theta(s)) + \text{log}P(\theta) = L_D + L_\theta \label{eq:medirl_loss}
\end{align}
\normalsize
Since $L_\theta$ can be optimized with weight regularization techniques for training \acp{NN}, \ac{MEDIRL} focuses on maximizing the first term $L_D$:

\vspace{-0.75\baselineskip}
\small
\begin{align}
    \cfrac{\partial L_D}{\partial\theta}
    = \cfrac{\partial L_D}{\partial R_\theta} \cfrac{\partial R_\theta}{\partial \theta}
    &= (\mu_D - \E[\mu]) \cfrac{\partial R_\theta(s)}{\partial \theta} \label{eq:medirl_update}
\end{align}
\normalsize
where $\E[\mu]$ is the expected SVF from the predicted reward.
In the \ac{MEDIRL} update equation \eqref{eq:medirl_update}, the derivative of the reward with respect to the weight parameter can be easily computed by back-propagation \cite{pytorch}.

\section{Costmap learning}
In this section, we introduce new \ac{IRL} algorithms for costmap learning.
We highlight that our proposed methods together provide more accurate and less expensive (without additional labeling) costmap compared to the original MEDIRL.

\subsection{Problem Definition}
We solve a trajectory planning problem of autonomous driving in a dense traffic highway scenario, where the main problem includes
\begin{itemize}
    \item An inference problem of a reward/cost function of the ego vehicle. Given an MDP without $R$, an observation $O_t(s_t)$, {goal $g$,} and an human demonstration data $s_{t,...,t+T}$, find {$R_\theta(s_t|g)$} that best explains $s_{t,...,t+T}$.
    \item A trajectory optimization problem. Given an MDP with $R=R_\theta(s_t|g)$, $s_t, O_t(s_t),$ and $g$, find the optimal path and control trajectory that maximizes $R$.
\end{itemize}

In the \ac{MDP} settings in our work, we use the term cost $(J)$ for \ac{OC}, and reward $(R)$ for \ac{IRL}, where one is defined to have the opposite meaning of the other. In $[0, 1]$ space, a reward can be represented as $R = 1 - J$, and vice versa.

The cost `map' we use in our work is the occupancy grid representation of a position state cost function, especially the ego-centric map having the ego vehicle at its center, always heading East.

\subsection{Costmap learning through \ac{IRL}}
\subsubsection{Settings}
Assumptions we make in this work are threefold. {(i)} The ego vehicle follows the kinematic bicycle model \cite{kong2015iv} shown below in \cref{eq:kinematic_bicycle}. {(ii)} We have a near-perfect state estimation of the ego and the surrounding vehicles within the ego's perception range. 
{(iii)} The driving demonstrations by human drivers show optimal behavior at an expert-level.

The discrete-time version of the kinematic bicycle model \cite{kong2015iv} we used for modeling our ego vehicle and computing the control actions for other baseline methods is written as:

\vspace{-0.75\baselineskip}
\small
\begin{align}
    \nonumber x_{k+1} &= x_k + v_k\text{cos}(\psi_k+\beta_k) \Delta t, ~~\psi_{k+1} = \psi_k + \frac{v_k}{l_r}\text{sin}(\beta_k) \Delta t  \\
    \nonumber y_{k+1} &= y_k + v_k\text{sin}(\psi_k+\beta_k) \Delta t, ~~v_{k+1} = v_k + a_k \Delta t \\
    \beta_k &= \text{tan}^{-1}\Bigg( \frac{l_r}{l_f+l_r} \text{tan}(\delta_k)\Bigg) \label{eq:kinematic_bicycle}
\end{align}
\normalsize
where $a$ and $\delta$ are the control inputs: acceleration and the front wheel steering angle. $\beta$ is the angle of the current velocity of the center of mass with respect to the longitudinal axis of the vehicle, $(x, y)$ are the position, the coordinates of the center of mass in an inertial frame $(X, Y)$, $\psi$ is the inertial heading angle, and $v$ is the vehicle speed. $l_r$ and $l_f$ are the distance from the center of the mass to the front and rear of the vehicle, respectively. The state $s$ is defined as $[x, y, \psi, v, \beta]$.

\begin{figure}[t]
\subfloat[\footnotesize Occupancy]{
  \centering
  \includegraphics[width=0.45\linewidth]{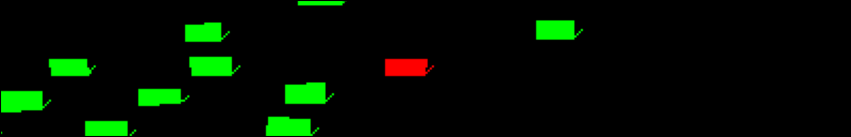}
  \label{fig:input:occupancy}
}
\subfloat[\footnotesize Velocity]{
  \centering
  \includegraphics[width=0.45\linewidth]{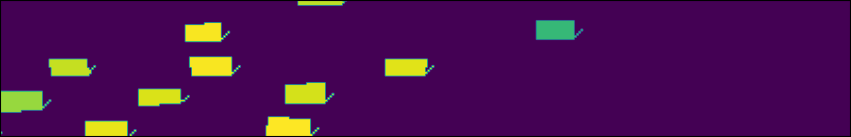}
  \label{fig:input:velocity}
}
\\
\subfloat[\footnotesize Acceleration]{
  \centering
  \includegraphics[width=0.45\linewidth]{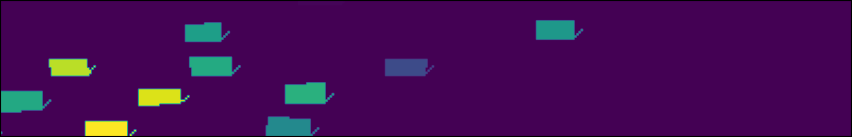}
  \label{fig:input:accel}
}
\subfloat[\footnotesize Heading angle]{
  \centering
  \includegraphics[width=0.45\linewidth]{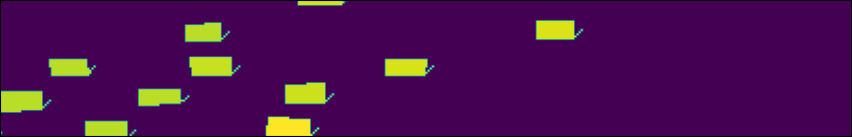}
  \label{fig:input:heading}
}
\\
\subfloat[\footnotesize Offset from the closest lane]{
  \centering
  \includegraphics[width=0.45\linewidth]{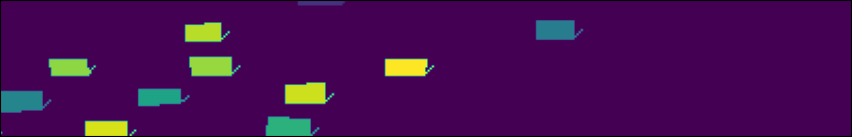}
  \label{fig:input:offset}
}
\subfloat[\footnotesize Source and goal lanes]{
  \centering
  \includegraphics[width=0.45\linewidth]{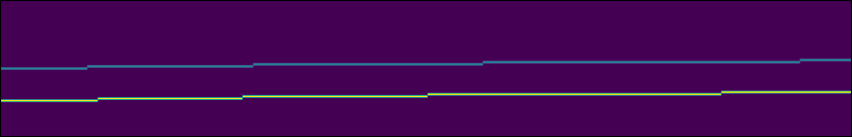}
  \label{fig:input:lanes}
}
\\
\centering
\subfloat[\empty]{
    \centering
    \includegraphics[width=0.45\linewidth]{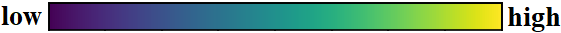}
}
\caption{\footnotesize The observations, {normalized and} converted into the bird's eye view ego-centric 2D images. The occupancy image shows two concatenated images of the ego (Red) and other vehicles' (Green) position information. The other images have their values on top of the occupancy map and brighter color represents higher values as shown in the color bar. In the lane image, the goal lane is brighter than the source lane.}
\label{fig:input}
\vspace{-\baselineskip}
\end{figure}

In our costmap learning, our reward NN model takes concatenated images (\cref{fig:input}) as input, where the images are composed of initial state features $f(s)$ obtained from raw sensor observations $O(s)$ at state $s$.
Starting from these initial raw state features, a better state features that better explain the reward are automatically extracted/learned implicitly inside the NN.

\subsubsection{Goal-conditioned \ac{IRL}}
The goal-conditioned \ac{RL} algorithms showed improvements in task performance in many \ac{RL} tasks {\cite{UVFA,Nasiriany2019goalconditioned}} compared to the regular \ac{RL} methods without a specified goal. Goal-conditioned policies learn $\pi_{\theta'}(a_t|s_t,g_t)$ instead of $\pi_{\theta'}(a_t|s_t)$. With the given goal information, the goal-conditioned planner focuses on which state to reach and improves the learning performance.
In our highway autonomous driving application, the source and the target lane information shown in \cref{fig:input:lanes} serves as a goal for lane changing and lane-keeping.
Especially for lane changing, without having this extra goal information, it is hard to make a prediction in a bimodal distribution case, where both left and right lanes are opened. 
Our goal-conditioned costmap learning avoids this multi-modal situation by specifying the goal and learning $R_\theta(s_t|g)$ instead of $R_\theta(s_t)$, with $g\in [\text{source, target}]$.

\subsubsection{Zeroing MEDIRL for an unexplored costmap}
\begin{figure}
\centering
\subfloat[\footnotesize A human demonstration of lane changing in bird's eye view.]{
  \qquad\qquad\qquad\includegraphics[width=0.5\linewidth]{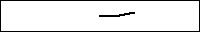}\qquad\qquad\qquad
  \label{fig:zeroout:demo}
}
\\
\centering
\subfloat[\footnotesize A costmap learned with the original MEDIRL.]{
  \qquad\qquad\includegraphics[width=0.5\linewidth]{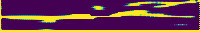}\qquad\qquad
  \label{fig:zeroout:original}
}
\\
\centering
\subfloat[\footnotesize A costmap learned with our Zeroing MEDIRL]{
  \qquad\qquad\includegraphics[width=0.5\linewidth]{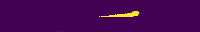}\qquad\qquad
  \label{fig:zeroout:ours}
}
\caption{\footnotesize A comparison example of the original \ac{MEDIRL} costmap and our zero learning method. (a): An example of the human demonstrations of lane changing depicted in a bird's eye view on 2D image space. The ego vehicle is located in the center of the image. A 3 seconds of a trajectory shown in the plot is used to learn a costmap. (b)-(c): The region with brighter color (yellow) represents lower cost region. Our method shows less noise and artifacts, thus provides less false positive errors to optimal controllers. A full comparison video can be found in the supplementary material.}
\label{fig:zeroout}
\vspace{-0.4cm}
\end{figure}

One major drawback of the previous approaches of \ac{MEDIRL} algorithms \cite{wulfmeier2015medirl, wulfmeier2016watchthis, wulfmeier2017largescale, Jung2021MEDIRL, zhang2018integrating} is that the learned cost representation includes a lot of artifacts and noise as shown in \cref{fig:zeroout:original}. The noise in the final cost map makes the \ac{RL} or \ac{OC} policy hard to find the optimal solution. Also, it is not intuitive to interpret the predicted cost map, because we don't know if the artifacts are false positive or not. The fundamental reason for this problem comes from the \ac{MEDIRL} algorithm itself, as described in the \ac{MEDIRL} paper \cite{wulfmeier2017largescale}, that the \ac{NN} model's weights are only updated in the visited area, i.e. the error feedback for unvisited states is not created and the network weights are not updated with respect to that error. To solve this real-world challenge of sparse feedback in the \ac{MEDIRL} algorithm, the \ac{MEDIRL} paper \cite{wulfmeier2017largescale} suggested using a pre-trained cost model which requires human labeling of reasonable cost features.

To solve the problem without requiring any human labeling or pretraining the model, we introduce a fully automated solution, the zero learning approach on top of the original \ac{MEDIRL} \cite{wulfmeier2017largescale} algorithm. As shown in \cref{fig:zeroout:ours}, the learned costmap excludes artifacts and noise for unvisited states and predicts high cost for the unvisited states. Predicting a high cost for the unknown and uncertain area is crucial for any safety-critical system. The resulting costmap with less noise and less false positive errors improves the performance of the optimal control by reducing the number of bad solutions coming from noise or errors in the cost.

We add another loss term, the zeroing loss, on top of the original \ac{MEDIRL} loss {in} \cref{eq:medirl_loss}:

\vspace{-0.75\baselineskip}
\small
\begin{align}
    L_{zero} = \ \sim(\mu_D + \E[\mu])
\end{align}
\normalsize
which minimizes the reward or maximizes the cost for `unvisited' states. The $(\sim)$ represents a NOT operator. We can think of this approach as supervised learning, having labels of 0 (low) reward for unvisited states, labeled by the demonstration and the learner's expected $\ac{SVF}$s. The total loss with this zeroing loss is defined as:

\vspace{-0.75\baselineskip}
\small
\begin{align}
    L(\theta) = L_D + L_\theta + c_{zero}L_{zero}
    \label{eq:zero_loss}
\end{align}
\normalsize
with a constant $c_{zero}$. Note that a big $c_{zero}$ could make the zeroing effect dominate other loss terms, resulting in predicting zero reward for all states. In practice, to balance with other losses, we choose $c_{zero}=T/(\text{cost map size})$, where $T$ is the number of demonstration timesteps in the costmap and the costmap size is its width $\times$ height.
The additional zeroing loss is minimized in a normal way of loss backpropagation as it has labels of 0 (reward) for unvisited states. Therefore, the update of $\theta$ with respect to the zeroing loss is independent to the original \ac{MEDIRL} update \cref{eq:medirl_update}.

\subsubsection{Spatiotemporal costmap learning}\label{sec:spatiotemporal}
\begin{figure}[t]
\centering
\subfloat[\footnotesize The optimal path (Red) with the original MEDIRL-learned non-temporal costmap.]{
  \centering
  \includegraphics[width=0.45\linewidth]{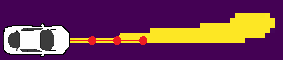}
  \label{fig:nontemp}
}
\hfill
\centering
\subfloat[\footnotesize The optimal path (Green) with spatiotemporal costmap, a concatenated image of (c)-(e).]{
  \centering
  \includegraphics[width=0.45\linewidth]{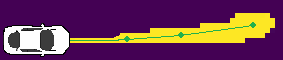}
  \label{fig:temp}
}
\\
\centering
\subfloat[\footnotesize Spatiotemporal costmap at $t=t_1$.]{
  \centering
  \includegraphics[width=0.28\linewidth]{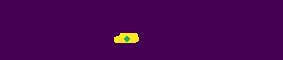}
  \label{fig:temp_t1}
}
\hfill
\centering
\subfloat[\footnotesize Spatiotemporal costmap at $t=t_2$.]{
  \centering
  \includegraphics[width=0.28\linewidth]{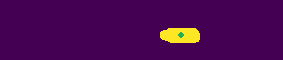}
  \label{fig:temp_t2}
}
\hfill
\centering
\subfloat[\footnotesize Spatiotemporal costmap at $t=t_3$.]{
  \centering
  \includegraphics[width=0.28\linewidth]{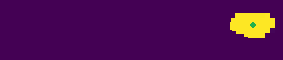}
  \label{fig:temp_t3}
}
\caption{\footnotesize A comparative illustration showing the optimal path found by the same optimal controller with and without the spatiotemporal learning costmap. 
Given only the costmap, the optimal controller is not able to find the optimal path for lane changing with the costmap without spatiotemporal learning, whereas with the spatiotemporal learning costmap, the lane changing path is obtained. Just for an illustrative purpose, the spatiotemporal costmap with $T=3$ is shown.
}
\label{fig:temp_nontemp}
\vspace{-0.4cm}
\end{figure}

The motivation of learning a spatiotemporal costmap is that the costmap obtained from the original \ac{MEDIRL} cannot be used by itself in MPC. Without temporal information, there are an infinite number of ways to follow the low cost region in the position costmap, many of which may cause collisions (\cref{fig:nontemp}).

In our spatio`temporal' costmap learning, the \ac{SVF} of each timestep's costmap is computed and used for updating $\theta$ in \cref{eq:medirl_update} {and results in costmaps that have temporal information, guiding which state to visit at which timestep, to get a low cost.}

We emphasize here that the Zeroing \ac{MEDIRL} algorithm we introduced above is necessary in practice for the spatiotemporal costmap learning because the quality of the spatiotemporal costmap is not acceptable without the Zeroing \ac{MEDIRL} as each costmap only predicts one timestep's costmap like in \cref{fig:temp_t1}, \cref{fig:temp_t2}, and \cref{fig:temp_t3}. It is vulnerable to noise and artifacts without the zero learning.

\subsubsection{Algorithm pipeline and \acl{NN} architecture}

\begin{figure}[b]
  \centering
  \includegraphics[width=\columnwidth]{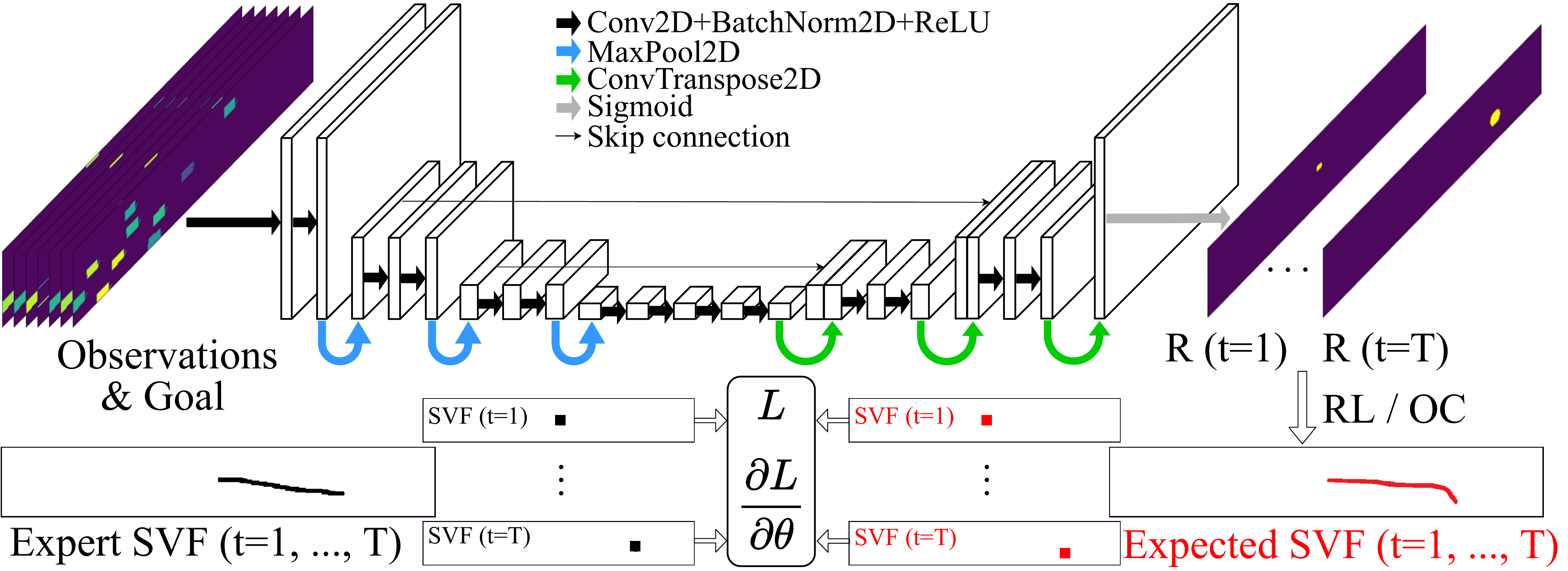}
\caption{\footnotesize The spatiotemporal costmap learning algorithm pipeline and a U-Net type \ac{NN} costmap model architecture with skip connections. The network takes the bird's eye view 2D image representations converted from observations and the goal information as input and outputs T costmaps, each representing each timestep's costmap. \ac{RL} or \ac{OC} policy finds an optimal path with the predicted costmap, and the \ac{SVFs} from the human demonstration and the policy are computed and used in $L$ and $\partial L / \partial \theta$ of the learning algorithm to update the \ac{NN} weights $\theta$.}
\label{fig:pipeline}
\end{figure}

\cref{fig:pipeline} shows the pipeline of our algorithm and the U-Net type \ac{NN} architecture we used to train the costmap model.
The reason why we use the image representation for the observation information is that, with the image representation, we can deal with a varying size of the number of neighbor vehicles within the ego vehicle's perception range \cite{Saxena2020RL}.
Given the observation {$O_t$} and the goal information {$g$}, our costmap model {takes their image representation as input and} predicts $T$ concatenated {position costmaps $J_\theta(x_{t+1},y_{t+1}|O_t,g),$} {$..., J_\theta(x_{t+T},y_{t+T}|O_t,g)$} at once, where $J_\theta$ is the position cost (map). The dimension of the output of the model is ($T$, width, height), where the width and height are 200 and 32 in the model we use, with a resolution of $0.5m$ per pixel.
Our optimal controller then finds optimal control and state trajectories with respect to the predicted costmap.
Next, from the MPC-propagated optimal states, we compute the \ac{SVF}s per each timestep. The \ac{SVF}s of the human demonstration are also computed and used in \cref{eq:medirl_update}, \eqref{eq:zero_loss} to update the weights of the \ac{NN} model.

\subsection{Optimal Control for solving the forward RL problem}\label{sec:optimal_control}

Given the reward from \ac{IRL}, we formulate the forward \ac{RL} problem in discrete time stochastic optimal control settings, where the vehicle model is stochastic, i.e. disturbed by the Brownian motion entering into the control channel, and we find an optimal control sequence $u^*$ in continuous action space such that{:}

\vspace{-0.75\baselineskip}
\small
\begin{align}
    u^*(\cdot) &= \underset{u(\cdot)}{\text{argmin}}  \E\Big[ \phi\big(s(T)|O_0,g\big)+\sum_{t=0}^{T-1} \pazocal{L}(s_t, u_t|O_0,g)\Big] \label{eq:stochastic_optimal_control}
\end{align}
\normalsize
where the expectation is taken with respect to dynamics \eqref{eq:kinematic_bicycle} with control $u$ having an additive Brownian noise $\mathcal{N}(0, \Sigma)$. Variable $s$ {denotes} the state $(x,y,\psi,v,\beta)$ defined in our vehicle model.

Since we will only use the position $(x,y)$-based costmap as a cost function to perform a task, we define $\pazocal{L}$ as:

\vspace{-0.75\baselineskip}
\small
{
\begin{align}
    \pazocal{L}(s_t, u_t|O_0,g) = \pazocal{L}(s_t|O_0,g) = J_\theta(x_t, y_t|O_0,g)
\end{align}
}
\normalsize
where {$J_\theta(x,y|O_0,g)$} is the {goal-conditioned} position costmap we learned through our \ac{MEDIRL} methods.
We define the final state cost {$\phi\big(s(T)|O_0,g\big)$} as

\vspace{-0.75\baselineskip}
\small
{
\begin{align}
    \phi\big(s(T)|O_0,g\big) = c_T J_\theta(x_T, y_T|O_0,g),
\end{align}
}
\normalsize
where $c_T$ is a constant value, set as 10.0 in our experiments.

While there are many approaches to solve \ac{RL} in autonomous driving \cite{isele2018safe,ma2020reinforcement,ppuu,Saxena2020RL}, to make full use of the known transition function we decided to use MPC. 
{Specifically, we apply a \ac{MPPI} controller \cite{mppi17} to solve the stochastic optimal control problem in \cref{eq:stochastic_optimal_control}.} 
While most MPC problems necessitate convexity and continuity of problems (and often closed-form solutions), \ac{MPPI} does not require a cost function or its derivatives to be convex, which fits our costmap formulation.

Following the information-theoretic derivation of the optimal control solution in \ac{MPPI} \cite{mppi17}, the control algorithm can be summarized as 1) sample a large number of Brownian noise $\mathcal{N}(0, \Sigma)$ sequence, 2) inject them to the control channels, 3) forward propagate the dynamics with the sequence of control + sampled noise (in parallel), 4) compute the cost defined in \cref{eq:stochastic_optimal_control}, 5) put more weights (exponentiated reward) on `good' noise sequences that resulted in a low cost, 6) update the control sequence with weighted noise sequence, 7) iterate the process until convergence. 8) execute the first $h$-timestep's control action. {Interested readers are referred to the original paper \cite{mppi17} for details.}

In theory, a large number of noise samples, a large noise ($\Sigma$), and a large number of iterations will result in an optimal solution, although the algorithm cannot be run in real-time. However, we use this ideal setting only when we do offline training, which does not require real-time performance.

\section{Spatiotemporal costmap \& MPC}
This section describes a variety of MPC problems that leverage the spatiotemporal costmap obtained by IRL.

\subsection{MPC with a learned costmap}
This approach uses the learned costmap in MPC.
The \ac{MPPI} described in \cref{sec:optimal_control} is used to optimize the given costmap, The number of samples and iterations are reduced to achieve a real-time path planning and control. We report this metod in \cref{tab:comparison} as MEDIRL-MPPI and GSTZ-MEDIRL-MPPI with their variants.

\subsection{Costmap as a path planner}
Another way of using the learned spatiotemporal costmap is directly using it as a path planner. We can extract waypoints from low-cost regions of each timestep's costmaps by finding the average positions {$(\bar{x}, \bar{y})$} of them. For example, the green circles in \cref{fig:temp} can be thought of as the waypoints extracted from the costmaps. These extracted waypoints can be used as an optimal path that a low-level trajectory tracking controller or MPC can follow.
We report this metod in \cref{tab:comparison} as GSTZ-MEDIRL-WPMPC with its variants.

\subsection{Quadratic programming with MPC}
The costmap-extracted average waypoints $(\bar{x}, \bar{y})$ are not smooth. More importantly, the waypoints are not physically constrained which may result in a low-level tracking controller failing to execute. Thus, we formulate a convex optimization problem with physical constraints based upon the vehicle dynamics \eqref{eq:kinematic_bicycle}. In particular, we incorporate $(\bar{x}, \bar{y})$ as state reference and, consequently, the problem is aligned with a formal reference tracking problem which a \ac{QP} solver \cite{cvxopt} is applicable. The convex problem reads:

\vspace{-0.75\baselineskip}
\small
\begin{align}
    \underset{u}{\text{min}}~ & \sum_{t=1}^{T} \cfrac{|| (x_t, y_t) - (\bar{x}_t, \bar{y}_t) ||_2}{T} \\
    \text{subject to } & \cref{eq:kinematic_bicycle}, ~ u \in [u_{\text{min}}, u_{\text{max}}], ~ \dot{u} \in [\dot{u}_{\text{min}}, \dot{u}_{\text{max}}] \nonumber
\end{align}
\normalsize
where the position state $(x, y)$ is a function of control $u=(\delta, a)$, shown in the kinematic bicycle model \eqref{eq:kinematic_bicycle}.

We recursively solve this problem in a receding horizon MPC fashion and in our paper, we call this MPC-QP method as WPMPC, which stands for WayPoint-following MPC.

\subsection{Recursive feasibility}
It is crucial for MPC that there exists a feasible solution at all times. Extracting a waypoint from the costmap does not guarantee a recursive solution.
We use a simple approach to guarantee the recursive feasibility of MPC with our spatiotemporal costmap:
At the $k$-th timestep, if the $k$-th waypoint is physically not reachable from $(k-1)$th waypoint with \cref{eq:kinematic_bicycle}, or if the $k$-th waypoint does not exist, we only use the waypoints up to the $(k-1)$th waypoints.

\subsection{Real-world limitation and an extra collision checker}
As shown in the next section, \cref{sec:experiments}, running MPC with a learned costmap does not always finish the task successfully (i.e. without collisions to other vehicles). Although the success rates are higher than 80$\%$ with our methods, there are several reasons our methods occasionally fail, which we discuss in the next section.
Since we solve a trajectory planning problem with a safety-critical system, it is required for the solution to be safe. For this practical reason, we add an extra safety-check pipeline on top of our \ac{IRL}-MPC framework.
The safety checker uses the same information, the other vehicles' state information, that we use to predict our cost and simply checks whether our MPC-predicted state trajectory will collide with other vehicles with some margin by simulating other vehicles for $T$ timesteps with a constant velocity model. If the collision checker detects a possible collision between the $k$-th ($k\leq T$) timestep's MPC-predicted ego states and the other vehicle states, the ego only executes $k-1$ steps of the MPC control sequence.

\section{Experiments}\label{sec:experiments}
We compare our proposed methods against multiple baselines described below.
Note that the Value Iteration policy in the original \ac{MEDIRL} algorithm was not used as a baseline since the optimal policy cannot be found and run in (5-20 Hz).
All the experiments were run with Nvidia GeForce GTX Titan XP Graphic Card and Intel Xeon(R) CPU E5-2643 v4 @ 3.40GHz $\times$ 12.
The models we trained use BatchNorm \cite{batchnorm}, PyTorch library \cite{pytorch}, and the Adam \cite{adam} optimizer.

\subsection{NGSIM data}
We collect lane change data from the Next Generation Simulation (NGSIM)\footnote{https://ops.fhwa.dot.gov/trafficanalysistools/ngsim.htm}, especially from the Interstate 80 Freeway dataset. We filtered and sorted out some noisy and unrealistic data and collected 240 lane-changing behaviors, divided into 215 train data and 25 test data.
The training data includes driving on the source lane with lane keeping, lane changing, and driving on the target lane with lane keeping after lane changing. 18 seconds in total per demonstration, 9 seconds before and after the middle of the lane change.
In this way, the model can learn the cost function that considers 1) determining the optimal state and timing to start changing lanes, 2) changing lanes, 3) keeping lanes before and after the lane change.
Each data point is a tuple $(O_t(s_t), s_{t,...,t+T})$, where $T=30$, 3 seconds of human demonstration with $dt=0.1$ in NGSIM data.
The observation space dimension we used in the experiments is (32, 200, 7), where 1 pixel represents $0.5m \times 0.5m$.

\subsection{Baseline methods}
\subsubsection{\acl{BC} (BC)}
For \ac{BC} policies, we used a ResNet \cite{he2016resnet} architecture and adjusted the dimesnions to match the our data (input) and the output (control dimension=2). As our width dimension is smaller than the input of the original ResNets, we also removed 2 max-pooling layers in the Fully Convolutional Networks. ResNets are one of the state-of-the-art \ac{NN} architectures with Convolutional \acp{NN} used for various types of image processing tasks and vision-based BCs \cite{Codevilla_2019_ICCV}.
We trained our custom ResNet models, ResNet 18, 34, and 50, with the same data we used to train our costmap model until the MSE loss with target actions converged to 0.02 and plateaued. Note that since the NGSIM data did not include the action information, we computed the actions from states following the bicycle kinematics \eqref{eq:kinematic_bicycle}.

\subsubsection{Model-free RL}
As described in \cref{sec:related_work}, PPUU \cite{ppuu} and DRLD \cite{Saxena2020RL} are the state-of-the-art policy learning methods that successfully demonstrated RL-based autonomous driving in dense highway traffic scenarios. Interested readers are referred to the original papers for details. In our experiments, we ran their pretrained models in our CARLA \cite{carla} testing environment described below. As they are not explicitly trained to do lane-changing, we report the success rate for these two methods computing the number of collision-free runs during the given task completion time.

\subsubsection{NNMPC}
The intention-based NNMPC \cite{bae2019cooperation} is a framework similar to ours, where it uses MPC to solve a path planning and control problem in a dense traffic highway scenario but solves a hard constrained optimization problem of obstacle avoidance with the predicted states of the other agents using Recurrent \acp{NN}. We use the original implementation of NNMPC \cite{bae2019cooperation}, and note that NNMPC and our method both use the same amount of information. While NNMPC directly solves a hard constrained optimization problem with a hand-designed and hand-tuned cost function, we solve the same path planning problem with a learned cost function.

\subsection{Simulation results and analysis}\label{sec:carla}
\begin{figure}
\subfloat[\footnotesize Our Goal-conditioned SpatioTemporal Zeroing (GSTZ) MEDIRL costmap shown in Red. The Red vehicle in the center is the ego vehicle, and the goal lane, source lane, and the other vehicles are in Green.]{
  \centering
  \includegraphics[width=0.55\linewidth,valign=t]{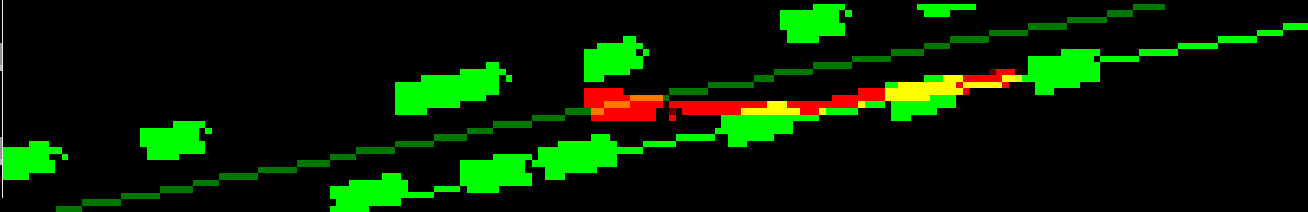}
  \label{fig:carla:obs_costmap}
}
\hfill
\subfloat[\footnotesize CARLA \cite{carla} used in our experiments.]{
  \centering
  \includegraphics[width=0.3\linewidth,valign=t]{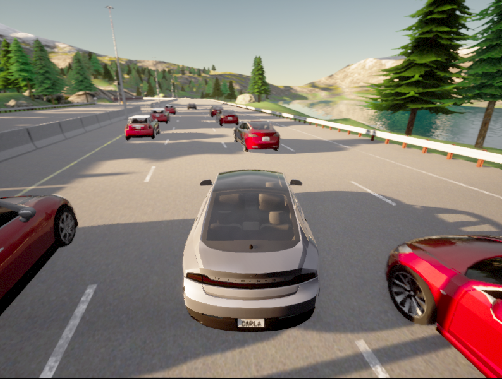}
  \label{fig:carla:scenario}
}
\caption{\footnotesize\textit{Left}: Our costmap (Red) shows a concatenated image of $T=30$ costmaps. \textit{Right}: The corresponding camera view from the CARLA \cite{carla} simulator showing our dense traffic highway lane changing scenario.}
\label{fig:carla}
\vspace{-\baselineskip}
\end{figure}

To test our proposed algorithms and the baselines with interacting agents, we ran experiments with the CARLA \cite{carla} simulator with ROS \cite{ros}. To reduce the gap between the real vehicle's model in CARLA and the kinematic bicycle model we use in MPC, we publish the MPC-predicted state trajectories as waypoints [$x, y, v$]. Then a low-level PID controller executes the vehicle's control commands (throttle and steering angle) to follow the MPC-generated waypoints.
\cref{fig:carla} shows our simulation environments and the performance of our costmap. 

\begin{table*}[t]
  \scriptsize
  \centering
  \begin{tabular}{l|p{0.5cm}p{0.5cm}p{0.5cm}p{0.5cm}p{0.7cm}p{0.5cm}p{0.5cm}p{0.7cm}p{0.5cm}p{0.5cm}p{0.5cm}p{0.5cm}p{0.5cm}}
  \toprule
  Model & Time & Succ. ($\%$) & Coll. ($\%$) & Time out ($\%$) & Brake Avg & Thr. Avg & Acc. Max & Brake Jerk Avg & Thr. Jerk Avg & Ang. Acc. Avg & Ang. Acc. Max & Ang. Jerk Avg & Ang. Jerk Max \\
  \midrule
  BC(ResNet18)
  & 14.45 & 44 & 56 & 0 & -0.34 & 0.63 & 1.55 & -0.59  & 0.68 & 0.24 & \cellcolor{green!25} 1.75 & \cellcolor{green!25} 0.45 & \cellcolor{green!25} 7.78 \\
  \midrule
  PPUU
  & 14.77 & 24 & 76 & 0 & -2.86 & 0.69 & 1.76 & -0.39 & 1.29 & \cellcolor{green!25} 0.17 & 3.24 & 0.52 & 27.49 \\
  \midrule
  DRLD
  & \cellcolor{green!25} 9.8 & 56 & 44 & 0 & -4.53 & 1.27 & 2.36 & -0.72 & 2.54 & 1.50 & 13.90 & 4.33 & 108.0 \\
  \midrule
  NNMPC
  & 13.20 & 86 & 14 & 0 & -0.74 & 0.54 & \cellcolor{green!25} 1.34 & -0.78 & 1.09 & 1.80 & 18.29 & 3.46 & 76.47 \\
  \midrule
  \specialcell[t]{MEDIRL-MPPI} &
  17.05 & 0 & 100 & 0 & -0.28 & \cellcolor{green!25} 0.53 & 1.73 & -0.48 & 0.55 & 0.50 & 4.62 & 0.90 & 9.39 \\
  \specialcell[t]{MEDIRL-MPPI-vel} &
  25.02 & 32 & 68 & 0 & \cellcolor{green!25} -0.17 & 0.55 & 2.09 & \cellcolor{green!25} -0.38 & \cellcolor{green!25} 0.43 & 1.70 & 16.36 & 3.16 & 39.14 \\
  \midrule
  \specialcell[t]{\bf{GSTZ-MEDIRL}\bf{-MPPI}} &
  31.88 & 74 & 10 & 16 & -0.37 & \cellcolor{green!25} 0.53 & 1.78 & -0.56 & 0.58 & 1.58 & 14.81 & 2.96 & 26.49 \\
  \specialcell[t]{\bf{GSTZ-MEDIRL}\bf{-MPPI-S}} &
  25.63 & 88 & 10 & 2 & -0.51 & 0.59 & 1.72 & -0.61 & 0.69 & 1.52 & 16.9 & 2.89 & 43.02 \\
  \midrule
  \specialcell[t]{\bf{GSTZ-MEDIRL}\bf{-WPMPC}}
  & 13.32 & 82 & 18 & 0 & -0.52 & 0.98 & 2.24 & -0.68 & 0.83 & 0.99 & 18.58 & 1.62 & 27.81 \\
  \specialcell[t]{\bf{GSTZ-MEDIRL}\bf{-WPMPC-S}}
  & 14.82 & \cellcolor{green!50} 96 & \cellcolor{green!50} 4 & 0 & -0.72 & 0.98 & 2.34 & -0.80 & 1.01 & 0.91 & 12.74 & 1.53 & 27.83 \\
  \bottomrule
  \end{tabular}
  \caption{\footnotesize Comparative analysis and ablation study (with N=50). The bold model names are our proposed methods. MEDIRL: a costmap learned with the original \ac{MEDIRL}. GSTZ-MEDIRL (ours): Goal-conditioned SpatioTemporal Zeroing MEDIRL. 
  MPPI: \ac{MPPI} controller. WPMPC: WayPoint-following MPC. -vel: with an additional velocity cost with a target velocity. -S: running an additional Safety layer of collision check on top of the IRL-MPC methods.
  }
  \label{tab:comparison}
\vspace{-0.4cm}
\end{table*}

\subsubsection{Scenario}\label{sec:scenario}
We designed a dense traffic highway scenario with 20 vehicles driving around the ego vehicle. Other vehicles perform lane keeping and collision avoidance and each vehicle tries to reach their target speed, which was randomly generated by $6 + U[-2,2]$ in $m/s$, where $U$ is uniform sampling.
{The behavior model of the other vehicles follows the Intelligent Driver Model (IDM) \cite{idm}, one of the well-known rule-based models, for lane following.}
The model is also based on the bicycle kinematics \eqref{eq:kinematic_bicycle}.
The other vehicles' behavior is designed to be always cooperative, where they slow down if the ego vehicle crosses a line in front of them and cuts into their lane.
We performed 50 experiments per algorithm where at each trial, the environment is randomized by starting with a different initial velocity of the ego vehicle and relative initial positions and target velocities of other vehicles.
For more details about the simulation environment and demonstration videos of our experiments, we guide the readers to check our supplementary video.

\subsubsection{Real-time Applicability of IRL-MPC}
The computation time of the whole pipeline (\cref{fig:pipeline}) majorly comes from the inference time of the costmap through NNs, which has $\mathcal{O}(1)$ time complexity, and solving the MPC problem. The time complexity of MPPI algorithm is $\mathcal{O}(n_iT)$, where $n_i$ is the number of iterations and $T$ is the number of timesteps. WPMPC, a QP solver with interior point methods, has $\mathcal{O}(n_v^3n_iT)$, where $n_v$ is the number of variables. With the computational resource described above, the computation time is approximately $100ms$, which was sufficient to match the ROS communication frequency with $10Hz$.

\subsubsection{Analysis}
We report in \cref{tab:comparison}, the performance of our proposed methods and the baseline algorithms.

First, we report that the \ac{BC} models were able to finish the task with about 80$\%$ of success rates with a simple scenario of other vehicles running in a constant speed maintaining a large constant gap. However, the \cref{tab:comparison} reports the results with a more challenging scenario we described in \cref{sec:scenario}. In this challenging scenario, \ac{BC} models were not able to finish the lane changing with more than 50$\%$ success {and we report the best model, ResNet18, in the table}.

As PPUU was trained with the entire NGSIM dataset that mostly includes driving straight and also because of the small clamping value for action ($0.01 rad$ for steering angle), it mostly drove straight until it crashes to the front vehicle. DRLD did a great job for both lane keeping and lane changing but the collisions happened mostly during lane changing. We believe these two results show that how fragile the RL-trained policies are when tested at a new environment.

The next baseline, NNMPC, was able to achieve 86$\%$ success.
Compared to the other baseline models, NNMPC does not only rely on the learned or trained models and it finds a rule-based optimal solution online on top of the NN-predicted behaviors.
Although the NNMPC has strict safety constraints in its optimization, we believe the prediction model of other vehicles might fail sometimes when other vehicles' velocity changed frequently.

Next, we ran \ac{MPPI} with the non-temporal costmap learned with the original \ac{MEDIRL} algorithm for comparison. Since we cannot extract correct waypoints from non-temporal costmap, we did not test the WPMPC with the non-temporal costmap.
As the non-temporal costmap does not include any notion of optimal velocity, unlike our spatiotemporal costmap, the MPPI starting with zero initial velocity does not find an optimal solution to do a lane change with the non-temporal costmap.
This behavior is shown in \cref{fig:nontemp}. However, once it explores the wrong/opposite direction to the goal lane, the costmap predicted at a new state (edge-case) is not correct, as the input data it takes has a very different distribution compared to the input in the training data.
We emphasize that this compounding error problem still exists in Deep \ac{IRL} and is one of the limitations of our cost function learning methods {that only learn from successful cases.}

We also tested MPPI with some initial velocity and the MEDIRL-learned non-temporal costmap with an extra velocity cost to maintain a target velocity at $10m/s$ with the MSE cost between the target and current velocities. {Although it showed a higher success rate compared to only using the original \ac{MEDIRL} costmap, it still reports a lot of failures.} Finding an optimal cost function that weighs between the two costs, position and velocity, is not an easy task, and even finding a good target velocity for accomplishing autonomous driving tasks is difficult. This {reminds} us of the main motivation of our spatio`temporal' costmap learning.

As expected, adding an extra safety check layer improved the success rates in all the models.
However, failures happened even with the safety check layer when the collision checker did not determine the collision would happen, based on the other vehicle's velocity.
Our future research will focus on improving our model to explicitly remove any potential collision-causing costmaps by itself, through a specific training procedure, so that it can achieve $100\%$ success rate without any extra safety-checker.

\subsubsection{Sensitivity Analysis on noise}
We also conducted the same experiments with a more realistic scenario by removing one of our assumptions of having a near-perfect state estimation. We injected an additive White Gaussian noise with different variance $\Sigma_N=[0.1, 0.1, 0.02, 1.0, 1.0]$ for different states $[x, y, \psi, v, acc]$, where $acc$ is the acceleration. The noise was added in the form of $c_s\cdot\epsilon$, with noise scale $c_s$ and $\epsilon\sim\mathcal{N}(0, \Sigma_N)$, to the estimated state of the other vehicles.
As shown in \cref{tab:sensitivity}, the performance degraded with bigger perception noise. From these experiments, we validated that there still exists a room for our method to improve, to make it more robust to real-world environments and to reduce the Sim2Real gap. We leave this direction for our future research.

\begin{table}[h!]
  \begin{center}
    \caption{Sensitivity Analysis of GSTZ-MEDIRL-WPMPC with Perception Noise}
    \label{tab:sensitivity}
    \begin{tabular}{l|c|c|c|c}
      \toprule
      \textbf{Noise scale $c_s$} & 0.0 & 1.0 & 2.0 & 5.0 \\
      \midrule
      \textbf{Success rate ($\%$)} & 82 & 80 & 74 & 68 \\
      \bottomrule
    \end{tabular}
  \end{center}
\vspace{-0.4cm}
\end{table}

\section{Conclusion}
In this work, we showed a new cost function learning algorithm that improves the original \acl{MEDIRL} (\ac{MEDIRL}) \cite{wulfmeier2017largescale} algorithm where our costmap can be directly used by MPC to accomplish a task without any hand-designing or hand-tuning of a cost function. Compared to the baseline methods, the proposed goal-conditioned spatiotemporal zeroing (GSTZ)-\ac{MEDIRL} framework shows higher success rates in autonomous driving, lane keeping, and lane changing in a challenging dense traffic highway scenario in the CARLA simulator.
We believe this work will serve as a stepping stone towards connecting \ac{IRL} and MPC.

\bibliographystyle{IEEEtran}

\bibliography{irlmpc}

\onecolumn
\section*{Citations}
\hspace{-0.3cm}Plain Text:
\\ \\
K. Lee, D. Isele, E. A. Theodorou, and S. Bae, “Spatiotemporal Costmap Inference for MPC via Deep Inverse Reinforcement Learning,” in IEEE Robotics and Automation Letters, 2022.
\\ \\
BibTeX:
\\ \\
@ARTICLE$\{$lee2022irlmpc,
\\author=$\{$Keuntaek $\{$Lee$\}$ and David $\{$Isele$\}$ and Evangelos A. $\{$Theodorou$\}$ and Sangjae $\{$Bae$\}$$\}$,
\\journal=$\{$IEEE Robotics and Automation Letters$\}$,
\\title=$\{$$\{$Spatiotemporal Costmap Inference for MPC via Deep Inverse Reinforcement Learning$\}$$\}$,
\\year=$\{$2022$\}$
\\$\}$

\end{document}